\documentclass[utf8]{article} % for Science, Engineering and Humanities and Social Sciences articles
\usepackage{url,hyperref,lineno,microtype,subcaption}
\usepackage[onehalfspacing]{setspace}

\usepackage{amsmath}
\usepackage{natbib}
\setcitestyle{round}
\usepackage{graphicx}
\usepackage[noend]{algpseudocode}
\usepackage[plain]{algorithm}
\usepackage{bbm}
\usepackage{colortbl}%for color in table
\providecommand{\keywords}[1]
{
  \small	
  \textbf{\textit{Keywords---}} #1
}

\newcommand{\footremember}[2]{%
    \footnote{#2}
    \newcounter{#1}
    \setcounter{#1}{\value{footnote}}%
}
\newcommand{\footrecall}[1]{%
    \footnotemark[\value{#1}]%
} 

\author{%
  Sophie Burkhardt\footremember{alley}{TU Kaiserslautern, Department of Computer Science, Kaiserslautern, Germany (work was performed while the author was at the University of Mainz)}%
  \and Jannis Brugger\footremember{trailer}{Johannes Gutenberg University of Mainz, Institute of Computer Science, Mainz, Germany }%
  \and Nicolas Wagner\footremember{darmstadt}{TU Darmstadt, Department of Computer Science, Darmstadt, Germany (work was performed while the author was at the University of Mainz)}%
  \and Zahra Ahmadi\footrecall{trailer}%
  \and Kristian Kersting\footremember{darmstadt2}{TU Darmstadt, Department of Computer Science and Centre for Cognitive Science, Darmstadt, Germany}%
  \and Stefan Kramer\footrecall{trailer}%
  }
\date{}

\begin{document}
%\onecolumn
%\firstpage{1}

\title{Rule Extraction from Binary Neural Networks with Convolutional Rules for Model Validation} 

% \author[\firstAuthorLast ]{\Authors} %This field will be automatically populated
%\address{} %This field will be automatically populated
%\correspondance{} %This field will be automatically populated

%\extraAuth{}% If there are more than 1 corresponding author, comment this line and uncomment the next one.
%\extraAuth{corresponding Author2 \\ Laboratory X2, Institute X2, Department X2, Organization X2, Street X2, City X2 , State XX2 (only USA, Canada and Australia), Zip Code2, X2 Country X2, email2@uni2.edu}

\maketitle

\begin{abstract}
Most deep neural networks are considered to be black boxes, meaning their output is hard to interpret. In contrast, logical expressions are considered to be more comprehensible since they use symbols that are semantically close to natural language instead of distributed representations. However, for high-dimensional input data such as images, the individual symbols, i.e. pixels, are not easily interpretable. We introduce the concept of first-order convolutional rules, which are logical rules that can be extracted using a convolutional neural network (CNN), and whose complexity depends on the size of the convolutional filter and not on the dimensionality of the input. Our approach is based on rule extraction from binary neural networks with stochastic local search. We show how to extract rules that are not necessarily short, but characteristic of the input, and easy to visualize. Our experiments show that the proposed approach is able to model the functionality of the neural network while at the same time producing interpretable logical rules.
\end{abstract}

\keywords{k-term DNF, stochastic local search, convolutional neural networks, logical rules, rule extraction, interpretability}

\section{Introduction}

\footnote{This paper is an extension of three workshop papers that were presented at the DeCoDeML workshops at ECML/PKDD 2019 and 2020: Sophie Burkhardt, Nicolas Wagner, Johannes F\"urnkranz, Stefan Kramer: Extracting Rules with Adaptable Complexity from Neural Networks using K-Term DNF Optimization; Nicolas Wagner, Sophie Burkhardt, Stefan Kramer: A Deep Convolutional DNF Learner; Sophie Burkhardt, Jannis Brugger, Zahra Ahmadi and Stefan Kramer: A Deep Convolutional DNF Learner.}Neural Networks (NNs) are commonly seen as black boxes, which makes their application in some areas still problematic (e.g., in safety-critical applications or applications in which DL is intended to support communication with a human user). Logical statements, however, are arguably much easier to grasp by humans than the main building blocks of NNs (e.g., nonlinearities, matrix multiplications, or convolutions). In general, learning of logical rules cannot be done using gradient-based algorithms as they are not differentiable. Even if we find rules that exactly describe a neural network, they might still be too complex to be understandable. To solve this problem, we propose {\textit convolutional rules} for which the complexity is not related to the dimensionality of the input but only to the dimensionality of the convolutional filters.

It is a wide-spread belief that shorter rules are usually better than longer rules, a principle known as Occam's razor. This common assumption was recently challenged again by \cite{Stecher:2016}, who revive the notion of so-called {\em characteristic rules} instead. As they show, shorter rules are often discriminative rules that help to differentiate different output classes, but are not necessarily descriptive of the data. In this work, we can confirm this observation and show how characteristic rules are produced for high-dimensional input data such as images.

Specifically, based on recent developments in deep learning with binary neural networks (BNNs) \citep{Hubara:2016}, we propose an algorithm for {\em decompositional rule extraction} \citep{Andrews:1995}, called {\em Deep Convolutional DNF Learner (DCDL)}.  A BNN takes binary input and only produces binary outputs in all hidden layers as well as the output layer. Some BNNs also restrict weights to be binary, however, this is not necessary for our approach. We then approximate each layer using rules and combine these rules into one rule to approximate the whole network. As our empirical results show, this allows for better approximation than with an approach that considers the neural network to be a black box --- a so-called {\em pedagogical rule extraction} approach. %#TODO Is DRS and DCDL the same?#

Moreover, we show how the convolutional rules can be used to visualize what the network has actually learned and that this visualization is more interpretable than visualizing the convolutional filters directly. %#TODO not sure about if the visualisation is more interpretable. 

To sum up, our contributions are as follows:
\begin{itemize}
    \item We formally define first-order convolutional rules (Section \ref{sec:first-order}) to describe a neural network using rules that are less complex than the original input.
    %\item (We compare the performance of our binary neural networks and the non-binary alternative to show that they are comparable.)
    \item We show that the decompositional rule extraction approach performs better than the approach that considers the network as a black box in terms of approximating the functionality of the neural network.
    \item We show how the convolutional rules produce characteristic visualizations of what the neural network has learned.
\end{itemize}

We proceed as follows. We start off by touching upon related work on binary neural networks, rule extraction, and interpretability in convolutional networks in Section \ref{sec:related}. Deep Convolutional DNF Learner (DCDL), consisting of the specification of first-order convolutional rules and the SLS algorithm, is then introduced in Section \ref{sec:method}. Our experimental results on similarity, accuracy,  and the visualization are presented and discussed in Section \ref{sec:results}.

\section{Related Work}
\label{sec:related}
Our work builds upon binary neural networks, rule extraction, and visualization of convolutional neural networks. 

\subsection{Binary Neural Networks}
Binary Neural Networks (BNNs) are neural networks that restrict the target of activation functions and weights to binary values $\{-1, 1\}$. The original motivation for BNNs is to reduce the memory footprint of NNs and accelerate inference \citep{Hubara:2016}. BNNs can be stored more efficiently because binary values can be stored in 1-bit instead of 32 bits or 64 bits. Also, binary representations can avoid computationally expensive floating-point operations by using less expensive, bitwise operations, leading to a speed-up at inference time \citep{Rastegari:2016}. 

However, by construction, the activation functions of BNNs lack differentiability and have less representational power due to their limitation to binary output values. Research on BNNs focuses on alleviating these two limitations.
A breakthrough for BNNs was the straight-through estimator (STE) introduced
in Hinton’s lectures \citep{Hinton:2012}. The STE calculates the gradient of the Heaviside step function $H$ as if it was the identity function. By using the STE in combination with the sign function $B(x) = 2 \cdot H(x) - 1$ instead of $H$, \cite{Hubara:2016} demonstrate the general capabilities of BNNs. They maintain real-valued weights while using binarized weights only for inference and calculation of gradients. Training updates are applied to the real-valued weights. They adapt the STE to better fit $B$ by clipping the identity function at -1 and 1 (Clipped STE). Nevertheless, the sole usage of the (Clipped) STE does not compensate for the lack of representational power of BNNs. Therefore, further improvements were proposed \citep{Rastegari:2016,Lin:2017}.
%More recent research incorporates short-cuts \citep{He:2016} which were originally intended to pass activations between non-consecutive convolutional layers. \cite{Liu:2018} use short-cuts to bypass the binarization of activations. This means, the output before binarization is used as an additional input to a subsequent convolutional layer. Thereby, they overcome the need for approximating real-valued activations through a linear combination of binary activations. Furthermore, they change the STE to be based on a piecewise polynomial function instead of the identity function (Poly. STE). In theory, this leads to a better gradient approximation \citep{Liu:2018}. In order to initialize a BNN, \cite{Liu:2018} use the parameters of a trained real-valued NN with the same architecture and the identity clipped at -1 and +1 as the activation function.

\subsection{Rule Extraction}
Rule extraction algorithms are commonly divided into decompositional and pedagogical approaches \citep{Andrews:1995}. Pedagogical (or model-agnostic) approaches view the neural network as a black box and approximate its global function using rules, whereas decompositional methods make use of the individual components of the network in order to construct the set of rules. Our work follows a decompositional approach, allowing a better approximation as compared to the pedagogical approach that we compare to in our experiments.

State-of-the-art pedagogical approaches include validity interval analysis (VIA), sampling, and reverse engineering. VIA \citep{Thrun:1993,Thrun:1995} searches for intervals in the input data within which the NN produces the same output. The found intervals can be transformed into rules. Approaches using sampling \citep{Craven:1995,Taha:1999,Sethi:2012,Schmitz:1999} try to let the NN label especially important instances in order to learn better rules. For instance, sampling can be beneficial to learn rules on parts of the unknown label function which are not covered well by the training instances. The reverse engineering approach by \cite{Augasta:2011} prunes the NN before the rules are extracted. As a result, the extracted rules are more comprehensible. \cite{Setiono:2000} use a similar technique to identify the relevant perceptrons of a NN.

Among others, decompositional algorithms use search techniques to find input combinations that activate a perceptron \citep{Fu:1994,Tsukimoto:2000}. Some search techniques provably run in polynomial time \citep{Tsukimoto:2000}. More recently, \cite{Zilke:2016} proposed an algorithm that extracts decision trees per layer which can be merged into one rule set for the complete NN. \cite{Gonzalez:2017} improve on this algorithm by polarizing real-valued activations and pruning weights through retraining. Both rely on the C4.5 \citep{quinlan:2014} decision tree algorithm for rule extraction. \cite{Kane:1993} use a similar idea but cannot retrain an arbitrary already existing NN. Right from the beginning, they train NNs having (almost) binary activations or perceptrons, which are only capable of representing logical AND, OR, or NOT operations. Rules are extracted by constructing truth tables per perceptron.

Unfortunately, the existing decompositional rule extraction algorithms have no principled theoretical foundations in computational complexity and computational learning theory. The runtime of the search algorithm developed by Tsukimoto \citep{Tsukimoto:2000} is a polynomial of the number of input variables. However, this holds only if the number of literals that constitute a term of an extracted logical rule is fixed. %#DELETE-BEGIN For the DCDL, the number of input variables of a rule extraction task can be up to 8192 in our experiments (see Section 5.4.1). Even if only 1\% of the input variables are used, this still leads to a polynomial of degree 82. #DELETE-END # This part is from the bachelor thesis and not part of the experiments. Probably we should read the original SLS paper again to describe the theoretical benefits of SLS better.Also from my experence In my experience, only the number of input variables alone is not a good criterion for evaluating the SLS. The diversity of the input is also important. Maybe I could make a series of experiments to determine the run time in practice and the accuracy for different k.#  

The other presented search algorithms exhibit an exponential runtime. Additionally, all mentioned search algorithms lack the possibility to fix the number of terms per extracted logical rule. Although the more recent decision tree-based approaches \citep{Zilke:2016} apply techniques to reduce the complexity of the extracted rules, they cannot predetermine the maximum model complexity. The only strict limitation is given by the maximum tree depth, which corresponds to the maximum number of literals per term of a logical rule. In general, the C4.5 algorithm does not take complexity restrictions into account while training. Directly being able to limit complexity, in particular the number of terms of a DNF in rule extraction, is desirable to fine-tune the level of granularity of a requested approximation.

In addition to pedagogical and decompositional approaches, there are approaches for local explanations, which explain a particular output, and visualization \cite{Ribeiro:2016}. Also, there are approaches that create new models that are assumed to be more interpretable than the neural network \citep{Odense:2019}. As this is not our goal, we focus on the decompositional approach in our work.

\subsection{Convolutional Networks and Interpretability}
Concerning the interpretability of convolutional neural networks, existing work can be divided into methods that merely visualize or analyze the trained convolutional filters \citep{Zeiler:2014,Simonyan:2013,Mahendran:2015,Zhou:2016} and methods that influence the filters during training in order to force the CNN to learn more interpretable representations \citep{Hu:2016,Stone:2017,Ross:2017}. Our work can be situated in between those two approaches. While we do change the training procedure by forcing the CNN to use binary inputs and generating binary outputs, we also visualize and analyze the filters after training by approximating the network with logical rules.

In order to make convolutional neural networks more interpretable, \cite{Zhang:2018} propose a method to learn more semantically meaningful filters. This method prevents filters from matching several different object parts (such as the head and the leg of a cat) and instead leads to each filter only detecting one specific object part (e.g., only the head), thus making the filters more interpretable. In contrast, our approach allows for different object parts being represented in one filter, but then uses the approximation with rules to differentiate between different object parts. One term in a k-term DNF (see Section \ref{sec:first-order} for the definition of k-term DNFs) might correspond to one specific object part. 
%Layerwise Knowledge Extraction from Deep Convolutional Networks

\section{Deep Convolutional DNF Learner (DCDL)}
\label{sec:method}

Our approach draws inspiration from recent work on binary neural networks, which are able to perform almost on par with non-binary networks in many cases \citep{Lin:2017,Liu:2018}. These networks are built of components that can provably be transformed into logical rules. The basic building blocks of any NN are variants of perceptrons. To ensure that a perceptron can be represented by a logical expression, we need to restrict the input as well as the output to binary values. This allows to transform perceptrons into truth tables. For hidden layers, we have to ensure that the output is binary leading to binary input for subsequent layers. For the input layer, we need to establish a binarization mechanism for categorical and a discretization mechanism for continuous features. The binarization of a categorical feature with $n$ possible values is done in a canonical way by expanding it into $n$ binary features. For the discretization of continuous features we use dithering. In particular, we use the Floyd-Steinberg algorithm\footnote{Implemented in \url{https://python-pillow.org/}.} to dither the gray scale images  to black and white images and dither the individual channels of RGB images. We tested Floyd-Steinberg, Atkinson, Jarvis-Judice-Ninke, Stucki, Burkes, Sierra-2-4a, and  Stevenson-Arce dithering\footnote{Based on \url{https://github.com/hbldh/hitherdither}, we also used the given error diffusion matrix.}, which are based on error diffusion, and found no statistically significant differences in the performance of the neural network using a corrected resampled t-test \citep{Nadeau:2003}.

A standard perceptron using the heaviside-function satisfies the requirement of binary outputs, but is not differentiable (unless using the delta-distribution). However, the straight-through estimator \citep{bengio:2013} calculates gradients by replacing the heaviside-function with the identity function and thus allows to backpropagate the gradients.

 %exploit the capabilities of the straight through estimator again. By setting the weights of an input perceptron to zero except for one input feature, the network itself conducts binarization through the perceptron's bias. We can add multiple input perceptrons per input feature for finer discretization. This mechanism also gives us control over complexity, as we can determine the number of bins. 

To be able to regularize complexity, we employ an adaptation of the stochastic local search (SLS) algorithm \citep{rueckert:2003} to extract logical expressions with $k$ terms in disjunctive normal form (k-term DNF). SLS can be parameterized with the number of terms to learn and thereby limit the maximum complexity. As the SLS algorithm is run after an NN has been trained, we do not limit the complexity at training time. \cite{know:2015} have already shown that this can be advantageous.

Convolutional Neural Networks (CNNs) are important architectures for deep neural networks \citep{dqn:2013,seq:2017,deepvar:2018}. Although convolutional layers can be seen as perceptrons with shared weights, logical expressions representing such layers need to be invariant to translation, too. However, logical expressions are in general fixed to particular features. To overcome this issue, we introduce a new class of logical expressions, which we call {\textit convolutional logical rules}. Those rules are described in relative positions and are not based on the absolute position of a feature. For inference, convolutional logical rules are moved through data in the same manner as convolutional filters. This ensures interpretability and lowers the dimensionality of extracted rules. %CNNs are the main motivation for using a decompositional approach. Parameters like stride and filter size can hardly be estimated in a pedagogical approach. 

Pooling layers are often used in conjunction with convolutional layers, and max-pooling layers guarantee binary outputs given binary inputs. Fortunately, binary max-pooling can easily be represented by logical expressions in which all input features are connected by a logical OR. The algorithms for training and testing DCDL are summarized in Algorithms \ref{alg:train_dcdl} and \ref{alg:test_dcdl}.%Skip connections \citep{skip:2015} are represented by identity functions.

\begin{algorithm}
\begin{algorithmic}[1]
    \Procedure{Train DCDL}{number of layers $L$} 
    \State $\phi\leftarrow \emptyset$ 
    \For{layer $l=1,\hdots,L$}
        %\State $\phi[l]$
        \If{Convolutional layer}
            \State $\psi \leftarrow \emptyset$
            \For{Convolutional filter $f$}
                \State $\psi_f\leftarrow$ SLS on input and output of NN for this filter
                \State $\psi\leftarrow \psi\cup \psi_f$
            \EndFor
            \State $\phi \leftarrow \phi\cup (l,\psi)$
        \ElsIf{Max pooling} 
            %\For{Output neuron}
                \State No training required
            %\EndFor
        \ElsIf{Dense}
            \State $\psi\leftarrow$ SLS on input and output of NN
            \State $\phi \leftarrow \phi\cup (l,\psi)$
        \EndIf
    \EndFor
    \Return{$\phi$}
    \EndProcedure
	\end{algorithmic}
	\caption{}
	\label{alg:train_dcdl}
	\end{algorithm}
	
	\begin{algorithm}
	\begin{algorithmic}[1]
    \Procedure{Test DCDL}{input data, trained SLS models $\phi$} 
    \State input$_1\leftarrow$ input data
    \For{$(l,\psi)\in\phi$}
        \If{Convolutional layer}
            \State $\lambda\leftarrow\emptyset$
            \For{rule $\psi_i\in \psi$}
                \State $\lambda_{i}\leftarrow $ evaluate rule $\psi_i$ on input$_l$
                \State $\lambda\leftarrow\lambda\cup \lambda_{i}$
%                 \If{$i=1$}
%                     output $\leftarrow \psi_i$
%                 \Else 
%                     $\, $output $\leftarrow$ output$\lor \psi_i$
%                 \EndIf
            \EndFor
        \ElsIf{Max pooling} 
            \For{Output neuron}
                \State Combine values in each pool with or-operation
            \EndFor
        \ElsIf{Dense}
            \State $\lambda\leftarrow$ evaluate $\psi$ on input$_l$
        \EndIf
        \State input$_{l+1}\leftarrow \lambda$
    \EndFor
    \Return{prediction of DCDL $\lambda$}
    \EndProcedure
	\end{algorithmic}
		\caption{}
\label{alg:test_dcdl}
	\end{algorithm}

\subsection{Introduction of First-Order Convolutional Rules}
\label{sec:first-order}

%Sophie will write this
This section provides the formal underpinnings and the introduction of the convolutional rules. We start with propositional k-DNF formulas and then move on to use first-order logic to take advantage of variable assignments (variables representing relative pixel positions) as we shift the filter across the image. 

A k-term DNF combines Boolean variables $\{x_0,x_1,\hdots,x_{n-1}\}$ as $k$ disjunctions of conjunctions
\begin{equation}
    \bigvee\limits_{i=0}^{k-1}\bigwedge\limits_{j=0}^{m_i}x_{i,j},
\end{equation}
where $x_{i,j}\in \{x_0,\neg x_0,x_1,\neg x_1,\hdots,x_{n-1},\neg x_{n-1}\}$ and $m_i\in \{0,1,\hdots,n-1\}$. An example with $k=3$ and 3 input variables could look like $(x_1\land x_2)\lor(x_3)\lor(\neg x_1\land \neg x_2)$. 

In general, a rule is described in relation to a fixed set of input variables. Unfortunately, for image data this is not sufficient. The success of CNNs in image classification arguably stems from the translation invariance of filters. Thus, we propose that logical rules for image classification need to be invariant to translation as well. In the following we assume all pixels are binary.

Using propositional logic, the definition of convolutional logical rules is relatively complex. For a first-order convolutional logical rule definition the rule itself is straightforward and the complexity is shifted to the definition of the predicates and the environment with respect to which the predicates are evaluated. Due to the variability of the environment that is inherent in first-order logic, we can naturally account for the translation invariance of the rule. In other words, the rule stays the same, only the mapping of the variables to the concrete values in the universe is changed as we move the rule over the image. Propositional logic on the other hand does not have variables and is thus not amenable to the translational invariance. 
% The following describes how such a rule can be defined for a 1-D input. For higher dimensional inputs such as images, the definition would be even more complex. Therefore, we also provide a definition using first-order logic, which leads to better readability.

% A k-term propositional convolutional logical rule (PCLR) $C$ that has $n$ Boolean input variables $\{x_0,x_1,\hdots,x_{n-1}\}$ is evaluated at $\mbox{floor}\left (\frac{n}{s}\right )$ positions which are placed in a 1-D grid over the input data. The stride $s$ determines the distance between subsequent evaluation positions. Thereby at maximum $t$ (window size) input variables can be part of a term. Formally, the output at the $j$th position is given by 
% \begin{equation}
%     o_j=\bigvee\limits_{m=0}^{k-1}\bigwedge\limits_{i=0}^{t-1}w_{C_{m,i}}\lor l_{js+i},
% \end{equation}
% where $w_{C_{m,i}}\in \{-1,1\}, 0\leq m<k,0\leq i < t$ are the shared parameters of $C$ and $l_{js+i}$ is equal to either $x_{js+i}$ or $\neg x_{js+i}$. If a shared parameter is equal to 1, the corresponding variable is not relevant. Padding with -1 is applied as necessary.

A k-term first-order convolutional logical rule (FCLR) is defined as follows: We define our model $\mathcal{M}$ as the tuple $(\mathcal{F},\mathcal{P})$ consisting of a set of functions $\mathcal{F}$ and a set of predicates $\mathcal{P}$ \citep{Huth:2004}. For a non-empty set $U$, the universe of concrete values, each predicate $P\in \mathcal{P}$ is a subset $P^{\mathcal{M}}\subseteq U^{a}$ of tuples over $U$, where $a$ is the number of arguments of predicate $P$. The universe $U$ of concrete values is defined as the set of concrete pixels in the image $p_1,\hdots,p_n\in \{0,1\}$. %#TODO add $P_conv$ in definition above?#%a set which is a union of two disjunct sets $U_0$ and $U_1$. $U_0:=\{u_{0,1},\hdots u_{0,n}\}$ is the universe of all pixels being zero and $U_1:=\{u_{1,1},\hdots u_{1,n}\}$ is the universe of all pixels being one and $U:=U_0\cup U_1$. The cardinality of the universe $|U|=2n$ is twice the number of pixels $n$. 

Now, a {\textit first-order convolutional logical rule} with one term is now defined as
\begin{equation}
 \phi_{conv}:=\exists x_1,\hdots,x_a: P_{conv}(x_1,\hdots,x_a),
 \label{eq:conv_rule}
\end{equation}
where $P_{conv}\in \mathcal{P}$ is the convolutional predicate and $a$ is the size of the convolutional filter or the number of elements in the filter matrix. The convolutional predicate is defined as 
\begin{equation}
    P_{conv}:=\{(u_1,\hdots,u_a)|u_i \mbox{ are consecutive pixels in accordance with the convolutional filter}\}.
    \label{eq:def}
\end{equation}
%$P_{conv}=\{(u_1,\hdots,u_m)|u_i\in U_0 \mbox{ if the } i \mbox{th pixel should be zero according to the convolutional filter and } u_i\in U_1 \mbox{ if the pixel should be one according to the convolutional filter, and } u_i\neq u_j \forall i,j.\}$. 

In order to evaluate our convolutional rule, we now need to specify the environment $l$ (the look-up table) \citep{Huth:2004} with respect to which our model satisfies (or not) the convolutional rule, i.e. $\mathcal{M}\models_l \phi$. The environment depends on the position at which we evaluate our convolutional rule. Evaluating the rule at position $t$ means that the variables $x_1,\hdots,x_a$ are mapped to the corresponding pixels in the input image,
$l_t[x_1\rightarrow p_{t}], l_t[x_2\rightarrow p_{t+1}],\hdots$, where $p_1,\hdots,p_n$ 
%$l_t[x_1\rightarrow u_{0,t}\mbox{ if }p_t=1\mbox{ else } u_{1,t}], l_t[x_2\rightarrow u_{0,t+1}\mbox{ if }p_{t+1}=1\mbox{ else } u_{1,t+1}],\hdots$, where $p_1,\hdots,p_n$ 
are the input image pixels (for 2D images, the indices have to be adjusted to account for the change to the next line, for simplicity, here the indices correspond only to 1D input). Thus, we are able to evaluate the convolutional rule in Equation \ref{eq:conv_rule} at each position of the image.

Since we specify the rules as $k$-term DNFs, we have $k$ convolutional predicates, one for each of the $k$ terms. Therefore, we can expand Equation \ref{eq:conv_rule} to
\begin{equation}
 \phi_{conv}:=\exists x_1,\hdots,x_a: P_{conv}^{(1)}(x_1,\hdots,x_a)\lor \hdots \lor  P_{conv}^{(k)}(x_1,\hdots,x_a).
 \label{eq:conv_rule_k}
\end{equation}
This concludes the definition of first-order convolutional rules.

\subsubsection{Example}
\label{sec:example}
\begin{figure}[t!]
	\centering
	\includegraphics[width=\textwidth]{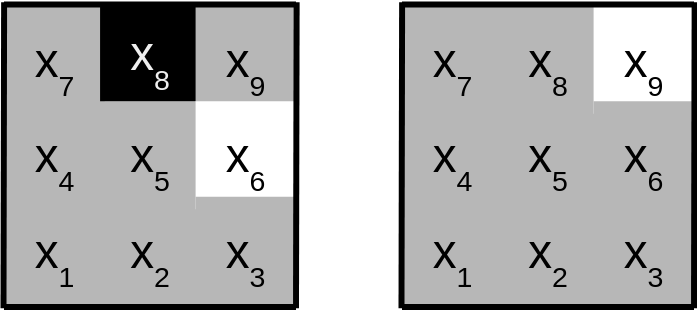}
		\caption{Graphical representation of the logical formula $P^{(1)}_{\mbox{conv}}(x)\lor P_{\mbox{conv}}^{(2)}(x)$, where $P_{\mbox{conv}}^{(1)}(x):=\{(x_1,\hdots,x_m)|x_6 \land  \neg x_8 \mbox{ is true}\} $ and $ P_{\mbox{conv}}^{(2)}:=\{(x_1,\hdots,x_m)| x_9 \mbox{ is true}\}$. The first convolutional predicate is displayed on the left side, the second one on the right side. Variables which have to be true in order for the predicate to evaluate to true are marked white. Variables which have to be false in order for the predicate to evaluate to true are marked black. Variables that have no influence on the evaluation are marked gray.} 
		\label{fig:graphical_representation_of_logical_formula}
\end{figure}

The logical rules found by SLS may be displayed graphically, if the input for SLS is image data. For each image position $t$ the variables of the convolutional predicates are mapped to the appropriate pixels using environment $l_t$. Thereby, each term of the k-term DNF can be visualized as an image. The whole convolutional rule can be output as a series of $k$ gray scale images by displaying positive literals as white, negative literals as black and literals that do not influence the truth value as gray. \autoref{fig:graphical_representation_of_logical_formula} shows an example of such a visualization with a rule that has two convolutional predicates and a filter size of $3\times 3$.\\
%Each pixel  $p_{ij}$ in the input image is mapped to $k$ many literals $x_{km}$ in the conjunct term (see Equation \ref{eq:def}). With $0<k< \mbox{maximum number of terms}$ and $m = \mbox{image width} * i + j$ (TODO: adjust in definition of conv rules to 2-D case).
%Each term of the k-term DNF can be visualized as an image by mapping the input image pixels to the variables in the convolutional rule (see Equation \ref{eq:def}) using an appropriate environment $l$.
%if $x_m$ not in conjunction term $\implies p_{ij} = 0$ \\??
%if $\neg x_m$  in conjunction term$\implies p_{ij} = -1$ \\??
%if $x_m$ in conjunction term$\implies p_{ij} = 1$ \\\\??

\subsection{Stochastic Local Search (SLS)}

We implemented the SLS rule learner\footnote{\url{https://github.com/kramerlab/DCDL}} and extended it for the purpose of pedagogical rule extraction (Algorithm \ref{alg:SLS}). Since we apply SLS to predictive tasks, we adjust SLS to return the candidate  that achieved the lowest score on the validation set (line 8). Scores used for the decision rule are still calculated on the training set. Calculation of scores is computationally expensive and SLS needs to evaluate the decision rule in every iteration. Therefore, we calculate scores batchwise. We introduce an adaptation that is theoretically motivated. One can always correct a term that falsely covers an instance by adding one literal, but the same does not hold in the case of an uncovered instance. We account for this by adjusting SLS to remove all literals in a term that differ from an instance (line 22).

Any SLS algorithm starts by evaluating a random solution candidate. It then selects the next candidate from a neighborhood of the former candidate. This procedure is repeated until a solution is found. If no solution is found and no improvement is found for 600 steps, we restart the search with a different random formula. Therefore, one has to define a candidate space, a scoring function to evaluate a candidate solution, a neighborhood of a candidate solution, as well as a decision rule for selecting the next candidate out of a neighborhood. 
In SLS, the candidate space consists of all applicable k-term DNFs. The scoring function is defined as the number of misclassified instances by a given k-term DNF. The neighborhood of a candidate is given by all k-term DNFs that differ in one literal to the candidate. The next candidate is selected in accordance with a randomly drawn misclassified training instance (line 12). If the instance has a positive training label, with probability $p_{g1}$ a random term is modified (line 14), otherwise the term which differs least from the misclassified instance. With the probability $p_{g2}$ the modification is done by deleting a random literal (line 20). In the case of a negative training label (line 25), any term that covers the considered instance is chosen. In contrast to before, a literal not in accordance with the misclassified instance is added with a probability of $p_{s}$. Otherwise, a literal whose addition decreases the score over the training set most is appended. In the end, SLS returns the candidate that achieves the lowest score on the training set.

\begin{algorithm}[t]
	\begin{algorithmic}[1]
    \Procedure{SLSearch}{$k, maxIteration, p_{g1}, p_{g2}, p_{s}, batchSize,  trainingStop, validationSet$} 
        \State $formula$ $\leftarrow$ a randomly generated $k$-term DNF formula
        \State $optimalFormula \leftarrow formula$
        \State $iteration \leftarrow 0$
       % \State $changedLast \leftarrow 0$
        \State $minScore \leftarrow \infty$
        %\State
        \While{ $iteration < maxIteration $ and $minScore > 0$}
        \State $iteration \leftarrow iteration +1$
            \State $newScore \leftarrow  score(validationSet)$
            \If{$newScore < minScore$}
               % \State $changedLast\leftarrow changedLast +1$
            %\Else
                \State $minScore \leftarrow newScore$
                \State $optimalFormula \leftarrow formula$
            \EndIf
            %\State $changedLast \leftarrow 0$ \textbf{if} $minScore$ has changed \textbf{else} increment by $1$
            \State $missedInstance \leftarrow$ random misclassified instance
            %\State
            \If{$missedInstance$ has positive label}
                \State \textbf{with probability} $p_{g1}$
                    \State\hspace{\algorithmicindent} $term \leftarrow$ a term uniformly drawn from $formula$
                \State \textbf{otherwise}
                    \State\hspace{\algorithmicindent} $term \leftarrow$ the term in $formula$ that differs in the smallest
                    \State\hspace{\algorithmicindent}\hspace{\algorithmicindent} number of literals from $missedInstance$

                %\State
                \State \textbf{with probability} $p_{g2}$
                    \State\hspace{\algorithmicindent} $literals \leftarrow$ a literal uniformly drawn from $term$
                \State \textbf{otherwise}
                    \State\hspace{\algorithmicindent} $literals \leftarrow$ all 
                    literals in $term$ that differ from \State\hspace{\algorithmicindent}\hspace{\algorithmicindent} $missedInstance$
      
               % \State
                \State $formula \leftarrow formula$ with $literals$ removed from $term$
               % \State
            \ElsIf{$missedInstance$ has negative label}
                 \State $term \leftarrow$ a term in $formula$ that covers $missedInstance$
                % \State

                 \State \textbf{with probability} $p_{s}$
                    \State\hspace{\algorithmicindent} $literal \leftarrow$ a literal uniformly drawn from all possibilities %\State\hspace{\algorithmicindent}\hspace{\algorithmicindent} so that $term \wedge literal$  does not cover $missedInstance$
                   
                 \State \textbf{otherwise}
                 \State\hspace{\algorithmicindent} $batch \leftarrow$ uniformly pick $batchSize$ many training instances
                    \State\hspace{\algorithmicindent} $literal \leftarrow$ a literal whose addition to $term$ reduces  \State\hspace{\algorithmicindent}\hspace{\algorithmicindent} $score(batch)$ the most

                % \State
                 \State $formula \leftarrow formula$ with $literal$ added to $term$
                % \State
            \EndIf
           % \EndIf
        \EndWhile
        \Return $optimalFormula$%a  $k$-term DNF formula that achieved $minScore$ 
       % \State
    \EndProcedure
	\end{algorithmic}
		\caption{}
		\label{alg:SLS}
\end{algorithm}

\section{Experimental Evaluation}
\label{sec:results}

The code for the following tests can be found on github\footnote{\url{https://github.com/kramerlab/DCDL}}. The parameters $p_{g1}$, $p_{g2}$ and $p_s$ are set to 0.5 for maximal randomness. We perform one-against-all testing so that the ground-truth-labels, which encode the classes of the data as one-hot vectors, are mapped to two classes. One class contains all images with the searched label, the other class contains all other labels. To prevent the neural network from predicting only the majority label, we balanced the labels in the training and test datasets such that one class comprises half of the dataset and the rest of the classes are randomly sampled so that each class is equally represented. We use three commonly used datasets with their predefined train-test splits: MNIST, FASHION-MNIST, and CIFAR10. For each dataset we used 5,000 samples as a holdout set for early stopping of the network and the rest for training. Each dataset has a designated test set with 10,000 samples.

\subsection{DCDL -- Similarity}
\begin{figure}[t!]	
    \centering
	\includegraphics[width=\textwidth]{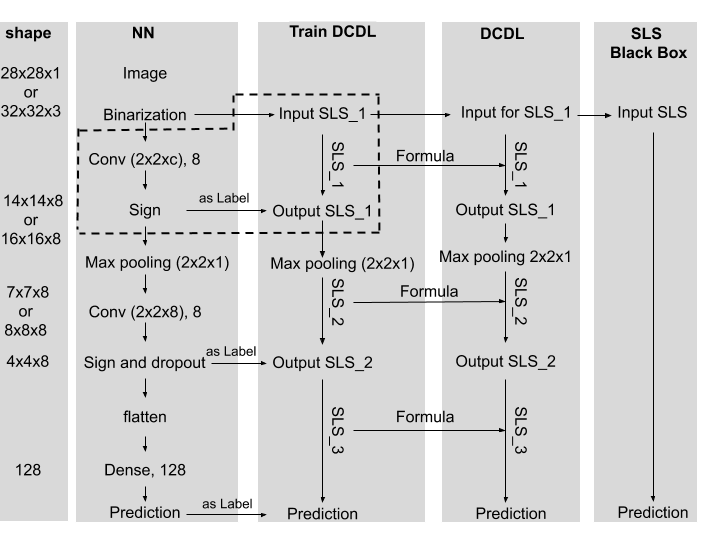}
		\caption{Experimental setup for comparing the neural network, the DCDL and the black-box SLS approach. $c$ depends on the dataset and is the number of color channels. The content of the dotted box is shown in more detail in Figure \ref{fig:aproximation_convolution}.}
		\label{fig:test_setup_accuracy_comparison}
\end{figure}

In this section we compare our DCDL approach against the black-box SLS algorithm.  We look at their ability to model the behavior of a multilayer neural network for the datasets MNIST, FASHION-MNIST and CIFAR10. An overview of the experimental setup is given in \autoref{fig:test_setup_accuracy_comparison}.

% #DELTE-BEGIN First a neural network is trained, which contains sign-layers for mapping values to -1 and 1. This is necessary to be able to apply the SLS algorithm in the next step. We use a network with small filters and few channels to keep the search space for the SLS algorithm small.
% The Mix column describes with which data the SLS algorithm is fed to find DNFs for approximating the neural network. The label is always extracted from the neural net. As input for the following SLS the output of the previous step is used instead of the data from the neural network because ....
% The Deep Rule Set column shows how the found DNFs are used successively for making a prediction #DELETE-END. 
%#ADD-BEGIN 
First a neural network is trained, which consists of two convolutional layers followed by a max pooling layer and a sign layer. The last layer is a dense layer with dropout. 
% #Answer in the 'big' net the last layer is a dense layer. In Figure 6 The dense Layer can be replaced by a simple function.  
As soon as the neural net is trained, the output of the sign layers is used as a label for the training of DCDL. The sign layers transform the outputs to binary values for the SLS algorithm. The two convolutional layers and the dense layer are each approximated with Boolean formulas, which are generated by the SLS algorithm. The dithered images are initially used as input to the SLS. After the first SLS, the input in the following SLS runs is the output of the previous formula. The intermediate results of the NN serve as labels. 

\begin{figure}[t!]
	\centering
	\includegraphics[width=\textwidth]{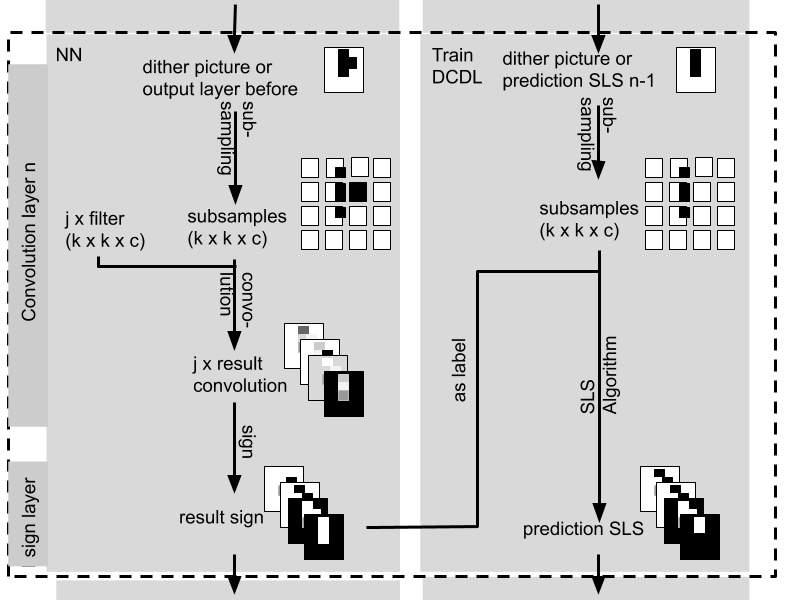}
		\caption{This figure illustrates the approximation of a convolutional layer by the SLS algorithm. It is part of the whole experimental setup in Figure \ref{fig:test_setup_accuracy_comparison} as shown by the dotted box.} 
		\label{fig:aproximation_convolution}
\end{figure}

The approximation of the convolutional operation is, in contrast to the dense layer, not straightforward, so we will explain this process in detail here. Figure \ref{fig:aproximation_convolution} shows the process graphically. 
In a convolutional layer, the input images are subsampled, and the samples are processed with the learned filters. Each sample is mapped to a value.  This mapping creates a new representation of the images. Each filter gives its own representation. They are stacked as different channels. With the help of the sign layer, the representations are mapped to binary values. 

The process of subsampling also takes place for the input of the DCDL approach. These samples are the input for the SLS algorithm. As labels serves the channel output of the sign layer belonging to the filter which is being approximated with the help of the SLS. Thus, each filter will be approximated by a logical formula. Using this procedure, DCDL approximates  the operation of the NN with Boolean formulas.

%#ADD-END

In the black-box SLS approach, only the input images and the corresponding label predicted by the NN are provided to the algorithm. The architecture of the NN and its functionality are not taken into account. It is evaluated using two different methods. In the black-box prediction approach, the prediction of the neural network is used as a label for training. In the true label approach, the true labels of the images are used for training.

\begin{figure}[htb]
	\centering
	\includegraphics[width=0.5\textwidth]{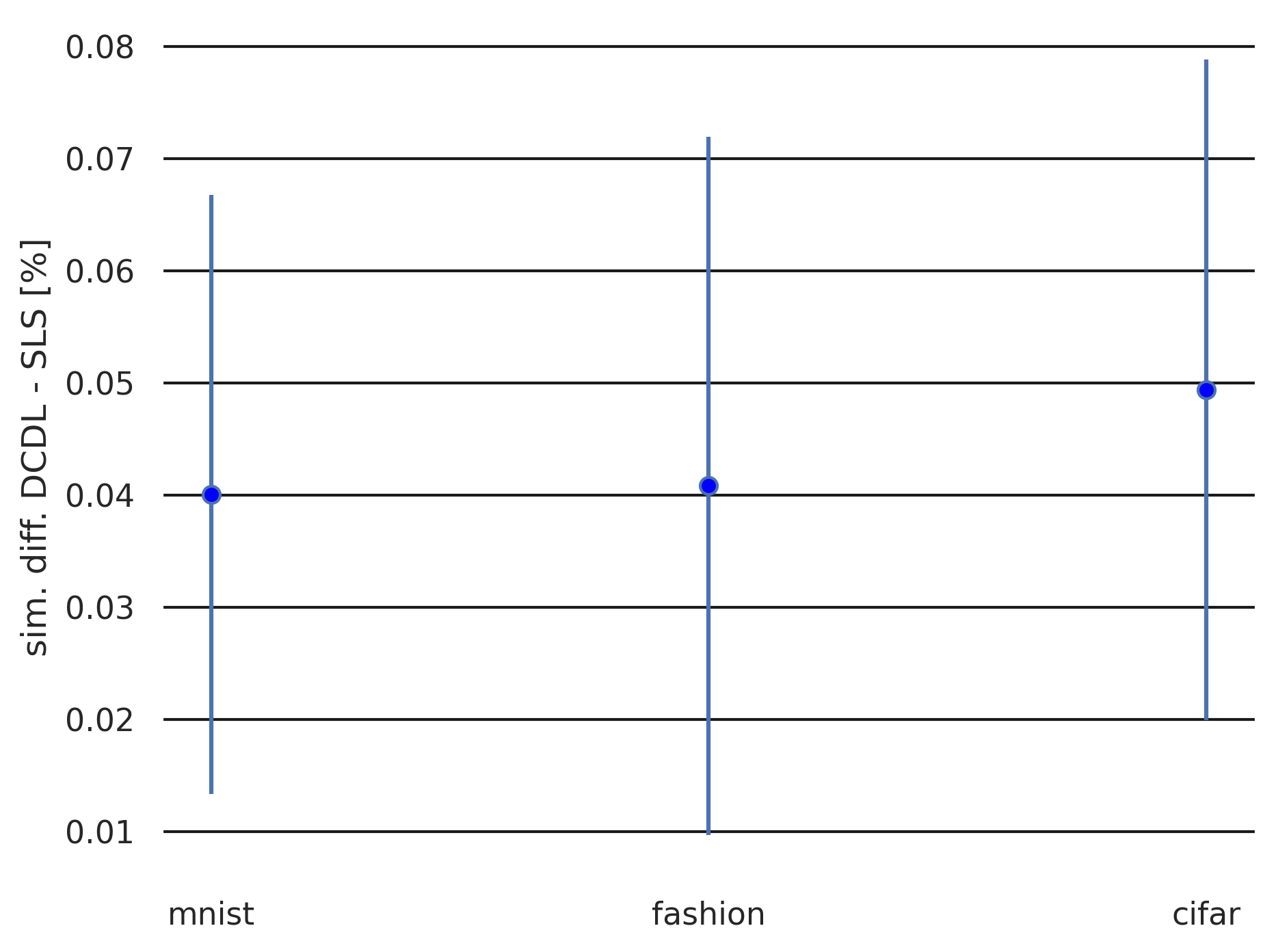}
		\caption{Difference in similarity between the prediction of the DCDL and the black-box approach for $k=40$. The similarity is calculated with respect to the prediction of the neural network on the test set. The DCDL approach has higher similarity across all three datasets. The standard deviation is calculated from 30 runs overall, 3 for each class.}
		\label{deep_rule_set_compare_black_box}
\end{figure}

We first focus on the question whether DCDL can better approximate the prediction of the neural network than the  black-box approaches.  Our results in \autoref{deep_rule_set_compare_black_box}  show that DCDL outperforms the black-box approach on the MNIST and Fashion-MNIST dataset and still provide a slight advantage on the CIFAR dataset. We hypothesize that the performance of the neural network itself has an influence on the ability of DCDL to extract good rules. CIFAR is a more complex dataset, and the neural network performs worse than on MNIST. This makes it also more difficult for DCDL to consistently model the network's behavior.

To calculate the similarity of the labels predicted by the neural network with the labels predicted by the SLS algorithm, we calculate
  \begin{equation}
       sim = \frac{ \sum_{i= 1 }^{ n } \mathbbm{1}[y'_i = y''_i]}{ n }
       \label{eq:equation_similarity}
  \end{equation}
   
with $n$ as the number of labels, $\mathbbm{1}$ the indicator function, $y'_i$ as the prediction of the investigated approach, and $y''_i$ as the label calculated by the neural network.%(Should we mention that the Factor of 2 comes because the label have the values -1 and 1?)\\

\subsection{DCDL -- Accuracy}

 The above section shows that our method performs well in terms of similarity with the neural network. However,  clearly, similarity does not necessarily correlate with accuracy. For example, it would be possible that the rule learner only models the errors that the network makes, leading to a high similarity but a bad performance on the actual labels. Therefore, we also compare the accuracy on the true labels of the predictions for the methods DCDL, black-box SLS, and the neural network. For the black-box SLS algorithm, we differentiate between the method that was trained on the labels as predicted by the NN (BB prediction) and the method that was trained on the true labels (BB label). Again we use \autoref{eq:equation_similarity} to calculate the accuracy,  but use the true labels of the test data instead of the labels predicted by the neural network. %(I use accuracy because now we use the real label) 

\begin{figure}[htb]
	\centering
	\includegraphics[width=\textwidth]{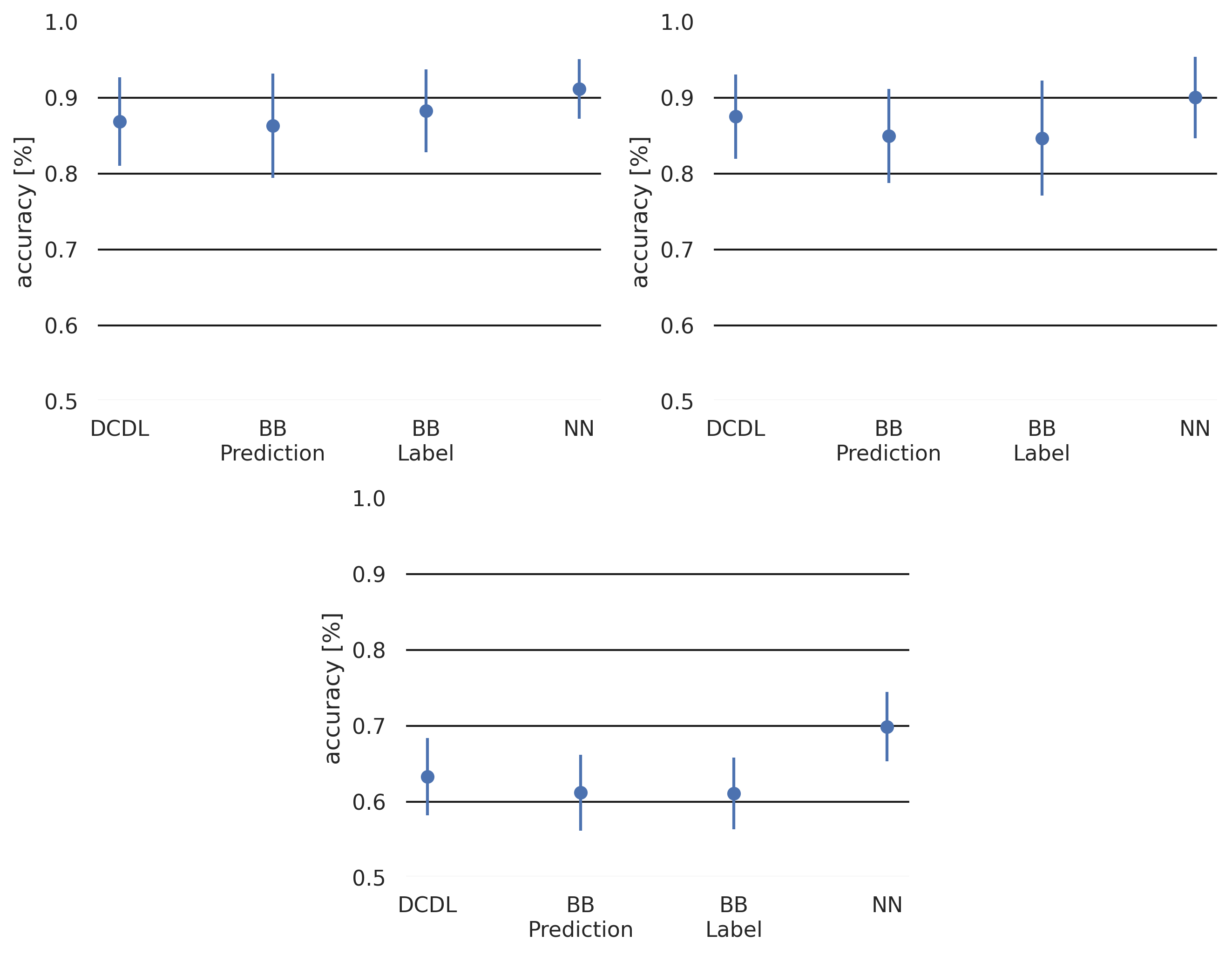}
	%#Add I added the figure created compltly with matplotlib should we 'zoom in' For the top row set y-axis limit to [0.7, 0.9] and for the lower row to [0.55 , 0.75]? old figure {figures/accuracy_all_models.png}
	%#TODO Add abbreviation to caption
		\caption{Accuracy for different datasets of our DCDL approach, the black-box SLS, black-box SLS with true labels, and the neural network. Top left: MNIST, top right: FASHION-MNIST, bottom: CIFAR. For SLS we used $k=40$. The standard deviation is calculated from 30 runs overall, 3 for each class.}
		\label{fig:accuracy_all_models}
\end{figure}

 Our results in \autoref{fig:accuracy_all_models}  show that the neural network outperforms the other methods. DCDL clearly outperforms the black-box approach on the MNIST and Fashion-MNIST datasets and still provides a slight advantage on the CIFAR dataset. The poor performance of the classifiers on the CIFAR dataset is most likely partly caused by the dithering. The network architecture might also play a role. This will be investigated further in our future work.
 
 As shown in Table \ref{tab:pvalue}, we also evaluated the statistical significance using a corrected resampled t-test \citep{Nadeau:2003} with $\alpha=0.05$ of the results to show that the performance of DCDL is similar to that of the neural network. For the MNIST and Fashion MNIST datasets, the neural network is not significantly better than DCDL, whereas the black box approach is significantly worse than the neural network. For CIFAR, all rule learning approaches perform worse than the neural network. The reason for this lies in the relative complexity of the dataset. We hypothesize that the network itself is not able to learn a good representation for the dataset, which in turn makes the task for the rule learners harder.

\setlength{\tabcolsep}{1.5em}
\begin{table}
\centering
\caption{{\em p}-values of the corrected resampled t-test \citep{Nadeau:2003} for MNIST, FASHION-MNIST and CIFAR for the accuracy values shown in Figure \ref{fig:accuracy_all_models}. Gray-shaded are the pairs for which the null hypothesis is rejected with significance level $\alpha = 0.05$ using a corrected resampled t-test \citep{Nadeau:2003}. }
\begin{tabular}{|l|c|c|c|} 
\hline
& BB prediction& BB label& Neural network~\\ 
\hline
\multicolumn{4}{|c|}{MNIST}\\                                                      
\hline
DCDL   & 0.59 & 0.27  & {\cellcolor[rgb]{0.753,0.753,0.753}}0.00 \\ 
\hline
BB prediction & -& 0.07 &{\cellcolor[rgb]{0.753,0.753,0.753}}0.00 \\ 
\hline
BB label & -  & - & {\cellcolor[rgb]{0.753,0.753,0.753}}0.02 \\ 
%\hline
%Neural network~ & - &- &- & -  \\ 
\hline
\multicolumn{4}{|c|}{Fashion}\\ 
\hline
DCDL& {\cellcolor[rgb]{0.753,0.753,0.753}}0.03 &0.10 & {\cellcolor[rgb]{0.753,0.753,0.753}}0.00                                     \\ 
\hline
BB prediction & -& 0.81&{\cellcolor[rgb]{0.753,0.753,0.753}}0.00  \\ 
\hline
BB label & -& -& {\cellcolor[rgb]{0.753,0.753,0.753}}0.01  \\ 
%\hline
%Neural network  & - & - &- & -\\ 
\hline
\multicolumn{4}{|c|}{Cifar}\\ 
\hline
DCDL& 0.08& {\cellcolor[rgb]{0.753,0.753,0.753}}0.04& {\cellcolor[rgb]{0.753,0.753,0.753}}0.00  \\ 
\hline
BB prediction & - & 0.90& {\cellcolor[rgb]{0.753,0.753,0.753}}0.00  \\ 
\hline
BB label&-& -                      & {\cellcolor[rgb]{0.753,0.753,0.753}}0.00  \\ 
%\hline
%Neural network  &-&- &- & -                       \\
\hline
\end{tabular}
\label{tab:pvalue}
\end{table}

\subsection{Visualization of logical formulas}

\begin{figure}[htb]
	\centering
	\includegraphics[width=\textwidth]{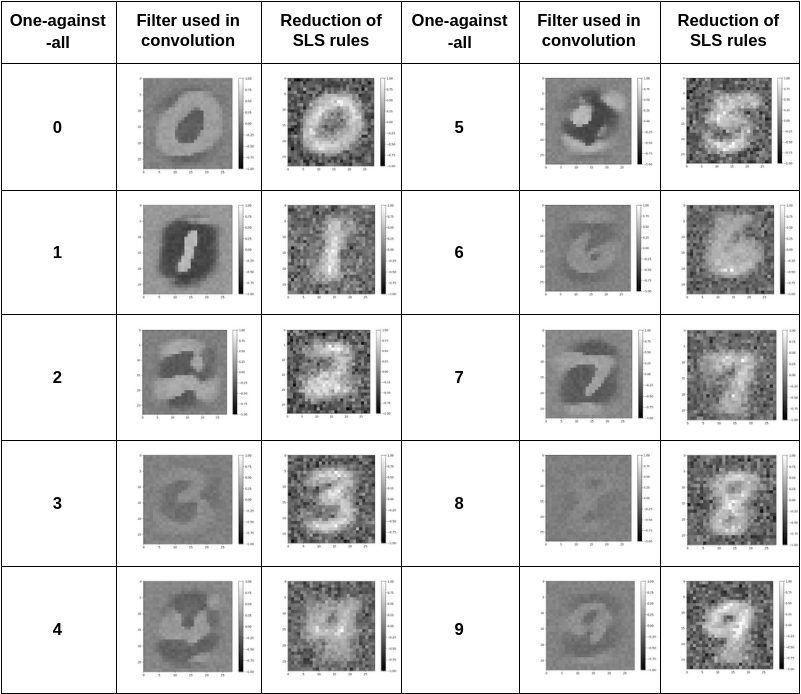}
		\caption{Results of visualizing convolutional rules rules with $k = 150$. The convolutional filter by the NN is shown for comparison.} 
		\label{fig:one_against_all_test_results}
\end{figure}

We already showed an example for the visualization of a simple formula in Section \ref{sec:example}. Now, we want to look at the visualization of more complex formulas that are found by our algorithm. If the rule search is conducted with a small $k$, the visualized rules tend to be discriminative and often highlight only a single pixel, thus making them hard to interpret. To counter this, we set $k$ to higher values in order to learn rules that are more characteristic. However, when visualizing $k$ predicates for high $k$, this produces a lot of images. Therefore, we add a reduce step that sums and renormalizes the visualizations for all $k$ predicates. As Figure \ref{fig:one_against_all_test_results} shows, this leads to clear visualizations that look almost like probability densities or prototypes. The comparison to the convolutional filters shows that this procedure leads to visualizations that are more easily identifiable by the human eye. 

\begin{figure}[t!]
    \centering
 	\includegraphics[width=\textwidth]{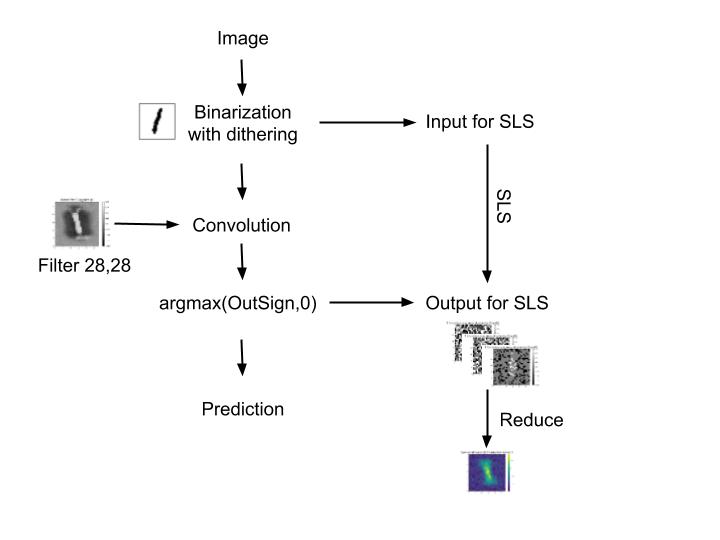}

		\caption{Test setup for approximating convolution operations with logical formulas.}
		\label{fig:test_setup_one_against_all}
\end{figure}

The architecture of the neural network is shown in \autoref{fig:test_setup_one_against_all}.  It consists of a convolutional layer followed by a sign layer and a dense layer. The dense layer converts the scalar output of the sign layer into a one-hot vector. The weights of the dense layer were set to [1,0].  %(We could add that in tensorflow the prediction of the NN is done by calling a argmax function on the prediction vector. Therefore the second value in the vector can be 0, because negative values are allowed for the first value. )\\\\
The dithered images are the input for the SLS algorithm and the output of the sign layer is the label for the SLS algorithm. The visualization in \autoref{fig:one_against_all_test_results} was done on the MNIST dataset with a filter size that is equal to the size of the image. Note that for the case of MNIST, smaller filter sizes do not result in interpretable visualizations. However, in principle we can also choose filter sizes much smaller than the image itself if the images consist of complex scenes where the number itself is only a small part of the image for example. The selection of a filter size that leads to an interpretable visualization is left for future work. To sum up, we showed with the help of a simple example how the individual predicates as well as the complete convolutional rule may be visualized.

%\section{Discussion}
%Sophie will write this when the results are finished

% #future work
% - visualitsation of hidden logic formulas
% #

\section{Conclusion}

We investigated how convolutional rules enable the extraction of interpretable rules from binary neural networks. We showed the successful visualization by means of an example. Additionally, the similarity to the functionality of the neural network was measured on three different datasets and found to be higher for the decompositional approach than the black-box approach. We think there is potential in decompositional approaches for the extraction and visualization of characteristic rules. Although the logical formulas are large for human visual inspection on real-world data, their representation makes deep learning models, in principle, amenable to formal verification and validation.

In future research, we aim to incorporate further state-of-the-art components of NNs while preserving the ability of the network to be transformed into (convolutional) logical rules. Our work suggests that the combination of binary NNs and k-DNF is promising combination. To this end, one should develop a differentiable version of DCDL based on, e.g., differentiable sub modular maximization \citep{Tschiatschek:2018} or differentiable circuit SAT \citep{Powers:2018}. Generally, one should explore DCDL as a new prespective on neuro-symblic AI \citep{Garcez:2020}. 

\section*{Permission to Reuse and Copyright}
Figures, tables, and images will be published under a Creative Commons CC-BY licence, and permission must be obtained for use of copyrighted material from other sources (including re-published/adapted/modified/partial figures and images from the internet). It is the responsibility of the authors to acquire the licenses, to follow any citation instructions requested by third-party rights holders, and cover any supplementary charges.
%%Figures, tables, and images will be published under a Creative Commons CC-BY licence and permission must be obtained for use of copyrighted material from other sources (including re-published/adapted/modified/partial figures and images from the internet). It is the responsibility of the authors to acquire the licenses, to follow any citation instructions requested by third-party rights holders, and cover any supplementary charges.

\section*{Conflict of Interest Statement}
%All financial, commercial or other relationships that might be perceived by the academic community as representing a potential conflict of interest must be disclosed. If no such relationship exists, authors will be asked to confirm the following statement: 

The authors declare that the research was conducted in the absence of any commercial or financial relationships that could be construed as a potential conflict of interest.

\section*{Author Contributions}

The experiments were done by JB. The initial code basis was due to NW, the writing of the paper was mainly done by SB with input from JB and NW. ZA and SK were involved in the development of ideas, the polishing of the paper, and discussions throughout. KK gave feedback to the paper and helped with the writing of the submitted manuscript.
%The Author Contributions section is mandatory for all articles, including articles by sole authors. If an appropriate statement is not provided on submission, a standard one will be inserted during the production process. The Author Contributions statement must describe the contributions of individual authors referred to by their initials and, in doing so, all authors agree to be accountable for the content of the work. Please see  \href{http://home.frontiersin.org/about/author-guidelines#AuthorandContributors}{here} for full authorship criteria.

\section*{Funding}
The work was funded by the RMU Initiative Funding for Research by the Rhine Main universities (Johannes Gutenberg University Mainz, Goethe University Frankfurt and TU Darmstadt) within the project ``RMU Network for Deep Continuous-Discrete Machine Learning (DeCoDeML)''.
%Details of all funding sources should be provided, including grant numbers if applicable. Please ensure to add all necessary funding information, as after publication this is no longer possible.

\section*{Acknowledgments}
%We thank Nicolas Wagner whose Bachelor thesis was the basis for this work, who wrote the Workshop manuscript for the DeCoDeML workshop 2019 and who provided invaluable support. 
Part of this research was conducted using the supercomputer Mogon offered by Johannes Gutenberg University Mainz (hpc.uni-mainz.de), which is a member of the AHRP (Alliance for High Performance Computing in Rhineland Palatinate,  www.ahrp.info) and the Gauss Alliance e.V.

\bibliography{Convolutional_rules_frontiers}

\end{document}